\documentclass{article} 
\usepackage{iclr2027_conference,times}

\usepackage{amsmath,amsfonts,bm}

\def\Figref#1{Figure~\ref{#1}}

\def\Secref#1{Section~\ref{#1}}

\def\eqref#1{equation~\ref{#1}}
\def\Eqref#1{Equation~\ref{#1}}

\def\1{\bm{1}}

\DeclareMathAlphabet{\mathsfit}{\encodingdefault}{\sfdefault}{m}{sl}
\SetMathAlphabet{\mathsfit}{bold}{\encodingdefault}{\sfdefault}{bx}{n}

\usepackage{hyperref}
\usepackage{url}
\usepackage{xspace}
\usepackage{booktabs}
\usepackage{graphicx}
\usepackage{amsmath}
\usepackage{multirow}
\usepackage{wrapfig}
\usepackage{subcaption}
\usepackage{titletoc}
\titlecontents{paragraph}[2.2em]{}{}{}{\titlerule*[0.5pc]{.}\contentspage}

\def\method{SplineWAM\xspace}

\def\Tabref#1{Table~\ref{#1}}

\title{\method: Adaptive Action Horizons for World Action Models via \mbox{B-Spline} \mbox{Representations}}

\author{Jun Guo$^{1,3}\thanks{Equal Contribution}$ , Xiaoshen Han$^{2,3*}$, Qiwei Li$^{3,4}$, Nan Sun$^{1,3}$, Peiyan Li$^{3,5}$, \\ \textbf{Heyun Wang$^{3}$, Hang Lai$^{3}$, Weinan Zhang$^2$, Xinghang Li$^{3}$\thanks{Corresponding Author} , Huaping Liu$^{1\dagger}$}
\\
\\
$^1$Tsinghua University, $^2$Shanghai Jiao Tong University, \\ $^3$Xiaomi Robotics, $^4$Peking University, $^5$CASIA
\\}

\iclrfinalcopy 
\begin{document}

\maketitle

\begin{abstract} World action models (WAMs) are large embodied policies that jointly predict future video and the actions to execute, emitting a fixed-length action chunk per inference call. Such a policy allocates its computational budget uniformly in time, unable to execute for longer over free-space motion or to spend more inference on contact-rich manipulation, which limits the throughput a WAM can reach when served in the cloud. We present \method, which adaptively compresses the action trajectory into a fixed-size window of cubic B-spline parameters, fitting the knot times to the characteristics of the motion. One parameter budget then decodes into chunks of varying temporal resolution and duration, and both the executed span and the interval until the next policy call follow from the prediction itself. Aligning the video supervision to the fitted knot times of the demonstration rather than to a uniform grid concentrates the supervised frames where the action trajectory is complex. For asynchronous deployment we introduce Jacobian-Pullback Real-Time Chunking (JP-RTC), which imposes chunk continuity on the decoded raw actions the robot executes rather than on the spline parameters, and corrects the parameters through the decoder so that the executed prefix agrees with the actions already committed. On LIBERO-Plus and RoboCasa, \method improves success rate over an action chunking WAM by $8.2$ and $4.4$ points while cutting policy calls per episode by $22\%$ and $26\%$. On three bimanual real-robot tasks under asynchronous execution, it leads or matches the baseline while decoding $1.2$ to $1.6$ times as much executed motion per call. Our project webpage is at: \url{https://splinewam.github.io/}.
\end{abstract}

\section{Introduction}

World action models (WAMs) have emerged as a practical way to transfer the dynamics knowledge of large video generative models into robot manipulation policies~\citep{Ye2026DreamZero,Ye2026GigaWorld,lingbotva,lingbotva2,Zhou2026Tau0WM,Yuan2026FastWAM}. By supervising future observations alongside future actions during training, a WAM learns not only which action to emit but also what that action does to the world. WAM designs differ widely in how the video is modelled and how tightly that model is coupled to action prediction, which has been explored thoroughly~\citep{Yuan2026FastWAM, Pai2025MimicVideo, Ye2026DreamZero}.

On the other hand, the way actions are represented and predicted has received less attention. Almost all current WAMs inherit the action chunk of \citet{Zhao2023ACT}: a fixed number of actions sampled on a uniform temporal grid. Action chunking is a strong default, and its benefits for behaviour cloning are by now well understood~\citep{Zhang2025ActionChunkingTheory}, but it ties three quantities together that need not be tied: the number of predicted parameters, the number of executed control steps, and the temporal resolution of the prediction.

\begin{figure}[t]
\begin{center}
\includegraphics[width=1.0\textwidth]{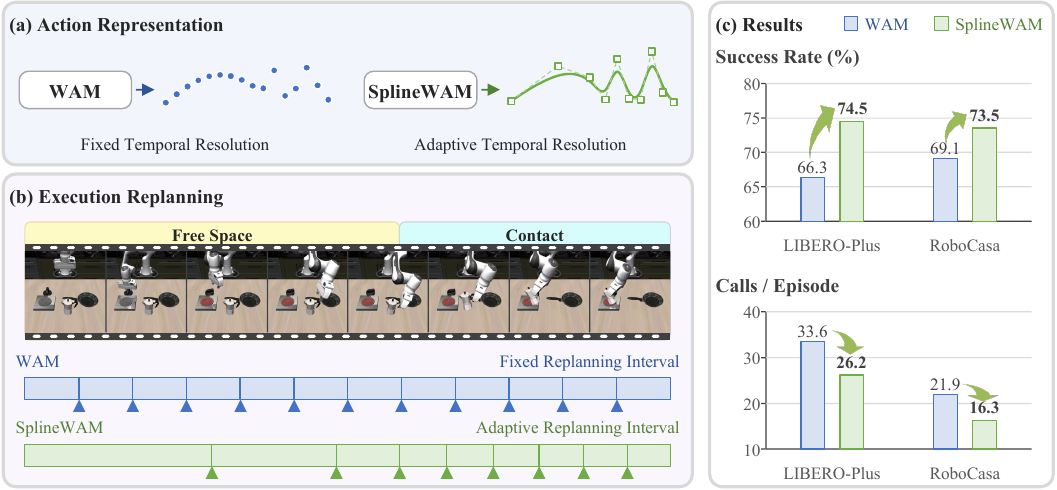}
\end{center}
\vspace{-0.1in}
\caption{(a) A WAM predicts a dense, uniform chunk of discrete actions, whereas \method predicts a continuous trajectory from a few adaptively spaced knots and their control points. (b) Those knots also set the replanning schedule, so chunks stay long through free space and shorten on contact. (c) The result is fewer policy calls per episode at a higher success rate.}
\label{fig:teaser}
\vspace{-0.2in}
\end{figure}

However, manipulation is not temporally uniform: reaching through free space is smooth, tolerant of error and predictable far ahead, whereas grasping, aligning and inserting need fine temporal resolution and can be committed to only briefly. A fixed chunk therefore spends more inference than free-space motion needs, and at the same time cannot replan promptly enough once contact begins. Both costs weigh heavily on an embodied model of this scale, since every policy call is a full forward pass of a video-pretrained backbone. A replanning interval too coarse for contact leaves the robot acting on increasingly stale observations, and redundant calls over free space consume inference for little task progress. Once such a policy is served to a fleet from shared cloud infrastructure, these redundant calls become a throughput bottleneck. A promising solution is to let the action representation carry its own horizon, so that the reach and the resolution of a prediction follow the motion rather than being fixed at design time.

To this end, we replace the action chunk of a WAM with a fixed-size window of B-spline parameters (\Figref{fig:teaser}). A B-spline describes a smooth curve through a list of \emph{knot times} and \emph{control points}. The policy predicts a fixed number of such pairs, and a decoder recovers executable actions from the spline at control steps. The number of decoded actions follows the predicted knot times, so one parameter budget covers a variable execution span, and with it a variable interval until the next policy call. Adaptive fitting also places knots densely where the trajectory is hard to approximate, so the temporal resolution of the prediction follows the complexity of the motion. The knot times further define the temporal axis of the video supervision. During training we sample video targets at the fitted knot times of the demonstration rather than on a uniform grid. Both branches are then supervised on one common non-uniform axis, and future frames receive supervision most densely where the action trajectory is most complex.

Deployment on a real robot additionally requires asynchronous execution, for which real-time chunking~\citep{Black2025RTC} offers a training-free solution with prefix guidance. Such a method is effective under action chunking, where the prefix is simply the leading entries of the generated variable, but in spline parameter space, where actions decode nonlinearly from the parameters, it is both inefficient and inflexible. We therefore propose Jacobian-pullback real-time chunking (JP-RTC), which instead forms the continuity residual on decoded actions and pulls it back through the decoder's local Jacobian.

Our contributions are threefold:
\begin{itemize}
\item We introduce a B-spline action representation into the action branch of a WAM, and align the video supervision to the fitted knot times of the demonstration. This fixes the parameter budget and lets the represented duration vary, so that a single prediction carries its own horizon and resolution.
\item We propose JP-RTC, which imposes chunk continuity on decoded actions and pulls it back through the decoder's local Jacobian, and thereby realise asynchronous, efficient and adaptive inference that is deployable on a real robot.
\item We evaluate on two simulated suites and on three bimanual real-robot tasks under asynchronous execution, ablating the representation of alignment and the fitting tolerance. \method gains $8.2$ and $4.4$ points of success rate over an action chunking WAM on the two suites while issuing $22\%$ and $26\%$ fewer policy calls per episode, and decodes $1.2\sim 1.6$ times as much executed motion per call on the real robot.
\end{itemize}

\section{Related Work}
\label{sec:related}

\paragraph{World action models and adaptive horizons.}
\label{sec:related:wam} Policies that predict future observations alongside future actions have been developed in diverse forms~\citep{Wu2023GR1,Ye2026DreamZero,Ye2026GigaWorld,lingbotva,lingbotva2,Guo2026XWAM,Zhou2026Tau0WM,Pai2025MimicVideo,Yuan2026FastWAM,Cai2026AHAWAM}, differing in how video is modelled and coupled to action prediction, but their action branches generally predict uniformly sampled, fixed-length chunks.

Future prediction makes WAMs expensive to run, and several works therefore target the cost of action inference. \citet{Yuan2026FastWAM} use an action-only test-time path, \citet{Guo2026XWAM} denoise video and action streams asynchronously, and \citet{Cai2026AHAWAM} decouple world prediction from action execution, adapting the horizon over which observation context is routed between the two while the action branch still emits a chunk on a fixed temporal grid. These approaches reduce the cost of producing a chunk, or vary how often it is conditioned, while retaining its fixed temporal support and sampling resolution. Outside the WAM literature, \citet{Liang2026AAC} select chunk sizes at inference time from predicted action entropy, while \citet{Pan2026VLACorrector} use visual discrepancy to trigger truncation and replanning. These approaches obtain adaptivity from auxiliary signals during inference; in \method, the horizon is encoded directly by the action representation before execution begins.

\paragraph{Action representations.}
\label{sec:related:action} Recent work represents actions with tokens derived from frequency-domain transforms~\citep{Pertsch2025FAST}, learned tokenizers~\citep{Liu2025FASTer,Dong2026ActionCodec}, prefix-decodable orderings~\citep{Liu2026OAT}, and quantized or hierarchical skills~\citep{Mete2024QueST,Lee2024VQBeT}; another line learns latent actions from action-free video for cross-embodiment transfer~\citep{Ye2024LAPA,Bu2025UniVLA,Tharwat2025LAWM,Chen2025villaX}. The temporal support of each represented segment remains fixed, while the number, hierarchy, or dimensionality of tokens varies. \method reverses this design: it fixes the parameter budget while allowing the represented duration to vary, which makes the action representation usable as an execution schedule.

\paragraph{B-spline actions.} \citet{Zhou2025BEAST} encode fixed-length action sequences as continuous or discrete spline tokens, while \citet{Lyu2025OmniSAT} use splines before residual quantization. \citet{Han2026BSP} parameterise actions by knots and control points, predict a fixed-size local window, and support time rescaling and resampling. We adopt its cubic representation and local-window construction, but use the predicted knot times to determine the execution horizon and replanning interval in an asynchronous WAM.

\section{Method}
\label{sec:method}

\subsection{Architecture}
\label{sec:architecture}

\method changes only what the action branch of a WAM predicts and leaves the rest of the model untouched, so any WAM that trains an action branch alongside a future-observation objective can adopt it. A sample consists of $L_{\text{video}}$ observation frames, encoded to video latents, and the B-spline action window of \Secref{sec:bspline} aligned to the current observation. Video latents and action parameters receive independent noise and independent flow-matching~\citep{lipman2022flow} timesteps, are embedded by their respective branches, and are then processed jointly under a shared attention mask. The action loss is defined on spline parameters, not on decoded actions. Appendix~\ref{app:impl} specifies the backbone, the training objective and its masking.

\subsection{B-spline actions}
\label{sec:bspline}

\paragraph{B-spline action curves.} Let a demonstration segment be a sequence of raw actions $\{(t_i, a_i)\}_{i=1}^{T}$ with $a_i \in \mathbb{R}^{d}$, where the time parameter $t_i$ is the integer frame index in the source episode. We represent the segment by a cubic B-spline curve of degree $k=3$,
\begin{equation} a(u) \;=\; \sum_{j=0}^{N_c} N_{j,3}(u)\, c_j ,
\label{eq:bspline}
\end{equation} with basis functions $N_{j,3}$ induced by a knot vector $U$ and control points $c_j \in \mathbb{R}^{d}$. All action dimensions share a single knot vector, so the temporal structure of the segment is described once rather than per dimension. We write the representation and its decoder as $z = [\,U, C\,]$ and $A = D(z)$, where $C$ collects the control points and $D$ evaluates \Eqref{eq:bspline} at integer control steps to recover executable actions. The representation is smooth by construction, has local support, and can be resampled at any control frequency. Because a WAM needs a fixed output shape, the policy predicts a fixed-size local slice of the episode spline rather than a whole trajectory; Appendix~\ref{app:impl} specifies how such a window is constructed and anchored to the current observation.

\paragraph{Fitting with adaptive knots.} How the knots are placed decides whether the representation can carry a horizon: only an error-driven fit lets the duration a window spans vary with the motion, so rather than spread knots uniformly over a fixed-length chunk at a fixed compression ratio as \citet{Zhou2025BEAST} do, we follow \citet{Han2026BSP} in fitting the whole episode trajectory under a bound on the reconstruction error, letting the number and spacing of knots follow the trajectory. Given a demonstration we fit \Eqref{eq:bspline} by least squares with an adaptive knot insertion loop in the style of FITPACK~\citep{dierckx1995curve}: fit with the current knots, insert a knot where the residual is largest, and repeat. Actions are normalised to $[-1, 1]$ over the dataset before fitting, and insertion halts once the largest elementwise deviation in that normalised space,
\begin{equation} E_{\max} \;=\; \max_{i,m}\,\bigl|\hat a_m(t_i) - a_m(t_i)\bigr| ,
\label{eq:emax}
\end{equation} taken over sample indices $i$ and action coordinates $m \in \{1, \ldots, d\}$, falls below a tolerance $\varepsilon$. Smooth stretches are then satisfied by few knots while high-curvature or contact-rich stretches attract many, so the knot distribution encodes local trajectory complexity and the temporal resolution it demands. The tolerance is the only knob of the representation and controls that resolution monotonically.

\subsection{Training: video--action temporal alignment}
\label{sec:training}

\begin{wrapfigure}{r}{0.42\textwidth}
\vspace{-\intextsep}
\begin{center}
\includegraphics[width=0.35\textwidth]{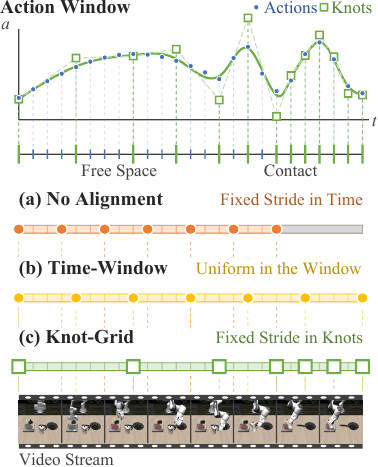}
\end{center}
\vspace{-0.1in}
\caption{Three ways to sample the video stream against one action window. The time axis above marks every action step, with the predicted knots among them.}
\label{fig:align}
\vspace{-0.2in}
\end{wrapfigure}

Because the action window carries explicit knot times, the video supervision no longer has to live on a grid of its own. \Figref{fig:align} contrasts the possibilities. A standard WAM subsamples the video stream at a fixed stride, which is set independently of the action window and can leave its later part unsupervised. Once the window supplies knot times there are two ways to place the two supervisions on a common time axis. \emph{Time-window alignment} keeps a uniform temporal grid over the local window, matching standard WAM practice. \emph{Knot-grid alignment} instead walks a fixed stride along the knots rather than along time: we take the future knot times from the window at that stride, prepend the current frame, and use the resulting frame offsets to select video targets, marking repeated tail knots as padding. These are the knot times of the offline fit of \Secref{sec:bspline}, cached with the demonstration. Knot-grid alignment makes the video supervision share the non-uniform temporal structure of the action supervision, so that frames are supervised most densely precisely where the action trajectory is complex, and the world model is asked to predict the future at the same resolution at which the policy must act there. All three variants supervise the same number of frames at the same subsampling ratio and differ only in which frames those are; \Secref{sec:exp:main} measures what the choice is worth.

\subsection{Inference}
\label{sec:inference}

\paragraph{Decoding a window into actions.} Decoding a predicted window at integer offsets $s = 0, 1, \ldots, \lfloor t_{\max} \rfloor$ yields $M = \lfloor t_{\max} \rfloor + 1$ actions, a count set by the predicted knot times and \emph{not} by the number of rows: at the same $L_{\mathrm{BSP}}$, widely spaced knots give a large $M$ and dense knots a small one. The decoded length is therefore a property of the prediction rather than a hyperparameter, and it determines how long the robot can act before the policy must run again, which is what adaptive horizons require. At execution time we give the decoder a budget of $H \le L_{\mathrm{BSP}}$ leading \emph{parameter rows}, the robot executes every control step they decode to, and the policy is called again when those steps run out. A call therefore commits to a slice of the parameterisation rather than to a step count, and the span it covers is whatever the knot times in that slice decode to, so at a fixed $H$ the executed span and the replanning interval vary between calls with the phase of the task.

\paragraph{Asynchronous execution.} Under asynchronous execution the robot keeps acting while the next window is being computed, so the prefix of the new prediction that overlaps the committed actions must agree with what is already being executed. Real-time chunking (RTC)~\citep{Black2025RTC} offers a training-free solution: it poses this as an inpainting problem and adds a pseudoinverse-guidance term to the sampler's velocity, formed from the residual between a mask-weighted prefix of the previous chunk and the one-step estimate of the chunk being generated, so that the new chunk is drawn towards agreement with the committed prefix.

For RTC the guidance target and the generated variable live in the same space, which for a per-timestep action chunk is also the space the robot executes. For a spline they are different spaces. The executed actions are $A = D(z)$ for a parameter window $z = [\,U, C\,]$, and $D$ is nonlinear in $z$: one control point influences several timesteps, neighbouring control points have overlapping support, and the knot times reparameterise the curve. A fixed number of parameter rows also spans a variable amount of real time, so a prefix defined by row count need not cover the actions the robot has committed to. Masking rows of $z$ therefore enforces parameter similarity when what asynchronous execution requires is agreement on executed actions.

\begin{figure}[t]
\begin{center}
\includegraphics[width=0.9\textwidth]{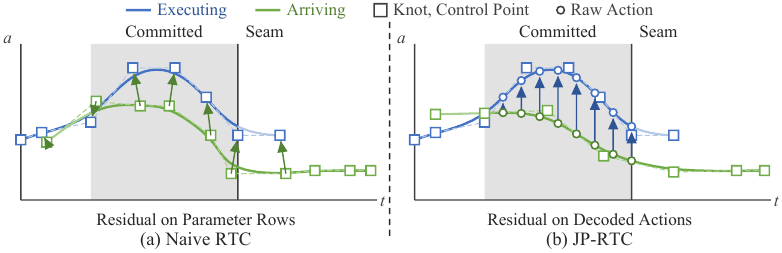}
\end{center}
\vspace{-0.1in}
\caption{Where the continuity constraint is imposed; arrows show what each rule pulls together. (a) Naive RTC must align parameter rows, including those outside the committed steps whose cubic support reaches into them. (b) JP-RTC constrains the decoded actions instead, so the two windows need neither matching rows nor equal knot counts.}
\label{fig:jprtc}
\vspace{-0.2in}
\end{figure}

\paragraph{Jacobian-pullback real-time chunking.} We impose the constraint where it is observed and let the generated variable follow; we call the result \emph{Jacobian-pullback real-time chunking} (JP-RTC); \Figref{fig:jprtc} contrasts the two choices. The residual is formed on decoded actions. We decode the prefix of the arriving window over a set $S$ of control steps and compare it against the actions the robot has committed to, namely the overlapping segment of the previous plan.

What remains is to convert a required change of actions into a change of parameters, and this is what the decoder's Jacobian $J_S$ supplies. It is the local linear map from a perturbation of the knot times and control points to the resulting displacement of the decoded prefix, so its transpose returns a correction expressed on actions to the parameters that would produce it. In doing so it accounts for the two properties that make the spline unsuitable for direct constraint: a single control point influences several control steps, and a knot displacement retimes the curve rather than translating it. Steering the sampler along the resulting direction gives the corrected velocity
\begin{equation} v_{\mathrm{JP}}(z^\tau_t, o_t, \tau) = v(z^\tau_t, o_t, \tau) + \min\!\left(\beta, \frac{1-\tau}{\tau \cdot r_\tau^2}\right) \delta^{\star}, \qquad \delta^{\star} = J_S^{\top}\bigl(J_S J_S^{\top} + \lambda I\bigr)^{-1} r_S,
\label{eq:rawrtc}
\end{equation} at flow-matching time $\tau$, where $r_S$ is the residual on decoded actions, weighted by a mask that decays over the overlap as in RTC, $J_S$ is evaluated at the sampler's current estimate of the clean window, and the guidance weight is that of RTC. The bracketed $\delta^\star$ is the parameter correction, obtained from a least-squares solve for the perturbation of the window that best cancels the residual, regularised by $\lambda$. A prefix of a few control steps is typically shorter than the interval a single knot span covers, so several parameters displace the decoded actions in nearly the same manner and the solve is genuinely underdetermined. The role of $\lambda$ is to select a correction within that ambiguity; Appendix~\ref{app:rtc} derives it, identifies it with the precision claimed for the match, sets its scale from $J_S$, and records the approximations the implementation makes. For a per-timestep chunk $J_S$ reduces to a selection of rows and \Eqref{eq:rawrtc} recovers the RTC update, so JP-RTC is a strict generalisation which departs from RTC only to the extent that the decoder departs from row selection.

Writing the residual after decoding also frees the two windows from having to be parameterised alike. JP-RTC never compares rows, so the arriving window is free to cover the committed steps with a different number of knots placed at different times and still be corrected to the same executed actions. What it requires is not a matching parameterisation but freedom in the rows that reach the constrained steps, and the regularisation resolves the residual ambiguity when those rows leave the solve underdetermined.

\section{Experiments}
\label{sec:experiments}

We ask whether a B-spline action window improves a WAM policy and where the gains come from (\Secref{sec:exp:main}), how much temporal compression is useful and how much of a prediction should be executed (\Secref{sec:exp:ablations}), and whether the representation holds up on a real robot (\Secref{sec:exp:real}).

\subsection{Setup}
\label{sec:exp:setup}

We report results on two simulated suites. \textbf{LIBERO-Plus}~\citep{fei2025libero} extends LIBERO~\citep{liu2023libero} into a robustness analysis of vision-language-action models by perturbing seven factors: background, camera pose, language instruction, lighting, object layout, initial state and sensor characteristics. \textbf{RoboCasa}~\citep{nasiriany2024robocasa} provides large-scale simulation of $24$ distinct kitchen tasks on a Franka Panda arm.

All methods share the Fast-WAM backbone~\citep{Yuan2026FastWAM}. We use the default fitting tolerance of each benchmark: $\varepsilon = 0.01$ on LIBERO-Plus, and $\varepsilon = 0.08$ on RoboCasa. Tolerances are absolute values in the normalised action space of \Secref{sec:bspline}, chosen per suite for comparable knot density. The executed row budget of \Secref{sec:inference} is $H = 12$ on LIBERO-Plus and $H = 20$ on RoboCasa, chosen from the sweep of \Secref{sec:exp:execlen}.

Baselines are \textbf{Action Chunking}, the unmodified Fast-WAM predicting a raw action chunk; \textbf{BEAST}~\citep{Zhou2025BEAST}, fixed-length B-spline tokenization of a fixed-duration sequence; \textbf{Naive B-Spline}, the representation and training recipe of \citet{Han2026BSP} without our alignment; and \textbf{Naive B-Spline w/o video}, the same variant trained with $\lambda_v = 0$ so that the auxiliary video term is removed. BEAST and Naive B-Spline denote the action representations of those papers ported into the Fast-WAM backbone, not their original models, and every baseline shares our backbone, demonstrations and training budget.

\subsection{Action representation and video--action alignment}
\label{sec:exp:main}

\begin{table}[t]
\caption{Success rate and inference efficiency by action representation, under synchronous execution at the default tolerance of each suite. Success is episode-weighted, Calls/ep.\ counts policy-model invocations, and Steps/call denotes the mean control steps executed per invocation. \method-Window and \method-Knot differ only in video--action alignment.}
\label{tab:main}
\begin{center}
\setlength{\tabcolsep}{4pt} \small
\begin{tabular*}{\textwidth}{@{\extracolsep{\fill}}lcccccc@{}}
\toprule
 & \multicolumn{3}{c}{LIBERO-Plus ($\varepsilon = 0.01$)}
 & \multicolumn{3}{c}{RoboCasa ($\varepsilon = 0.08$)} \\
\cmidrule(lr){2-4}\cmidrule(l){5-7}
Action representation & Success & Calls/ep. & Steps/call
& Success & Calls/ep. & Steps/call \\
\midrule
Action Chunking & 66.3 & 33.6 & 8.00 & 69.1 & 21.9 & 16.00 \\
BEAST & 66.6 & 35.5 & 8.00 & 65.0 & 24.6 & 16.00 \\
Naive B-Spline w/o video & 60.0 & 38.7 & 7.92 & 55.0 & 32.7 & 14.26 \\
Naive B-Spline & 70.7 & 27.3 & 8.36 & 69.5 & 16.9 & 17.98 \\
\midrule
\method-Window & 73.3 & 27.1 & 8.39 & 71.9 & 17.5 & 18.10 \\
\method-Knot & \textbf{74.5} & \textbf{26.2} & \textbf{8.41} & \textbf{73.5} & \textbf{16.3} & \textbf{18.18} \\
\bottomrule
\end{tabular*}
\end{center}
\vspace{-0.2in}
\end{table}

In \Tabref{tab:main}, \method-Knot improves on the action chunking WAM by $8.2$ points on LIBERO-Plus and $4.4$ on RoboCasa while issuing $22\%$ and $26\%$ fewer policy calls per episode, and knot-grid alignment is the better of the two alignments on both suites, by $1.2$ and $1.6$ points.

BEAST does not benefit: it applies spline tokenization while keeping the duration of a token block fixed, and lands $0.3$ points above the baseline on LIBERO-Plus while falling $4.1$ points below it on RoboCasa. A spline parameterisation alone therefore does not account for the gains, since BEAST supplies one and does not obtain them; what separates it from the other spline rows is that the duration it represents is fixed in advance.

The two Naive B-Spline rows differ only in whether the auxiliary video term is present, and removing it costs $10.7$ and $14.5$ points. Joint video--action training is therefore doing substantial work for the action branch, which agrees with the finding of \citet{Yuan2026FastWAM}.

\subsection{Ablations}
\label{sec:exp:ablations}

The comparison of \Secref{sec:exp:main} evaluates the representation at a single operating point, which is determined by two quantities. We ablate each of them in turn. The fitting tolerance $\varepsilon$ sets how densely knots are placed in the training data, a property of the dataset the policy is fitted to; the row budget $H$ is chosen at deployment time on a trained model.

\begin{figure}[t]
\begin{center}
\includegraphics[width=\textwidth]{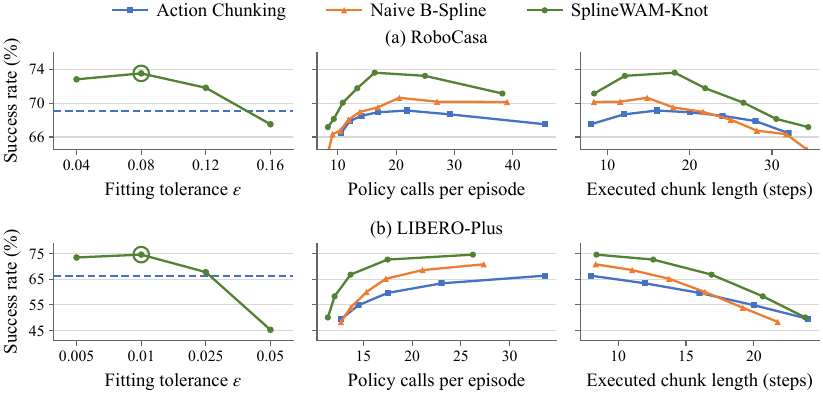}
\end{center}
\vspace{-0.1in}
\caption{Ablations on (a) RoboCasa and (b) LIBERO-Plus: success rate against the fitting tolerance $\varepsilon$ (left), and against the two costs of the executed length (middle, right). Dashed line is the action chunking baseline; the circled point is the tolerance we deploy.}
\label{fig:ablations}
\vspace{-0.1in}
\end{figure}

\paragraph{Fitting tolerance.}
\label{sec:exp:inference} The left panel of each row of \Figref{fig:ablations} sweeps the fitting tolerance from a setting tight enough to track the demonstrations almost point by point to one loose enough to discard most knots. Each point is a separate \method-Knot model trained under the identical budget on demonstrations refitted at that tolerance, $\varepsilon$ being a preprocessing parameter the model is never conditioned on.

\begin{figure}[t]
\begin{center}
\begin{subfigure}[t]{0.32\textwidth}
  \includegraphics[width=\textwidth]{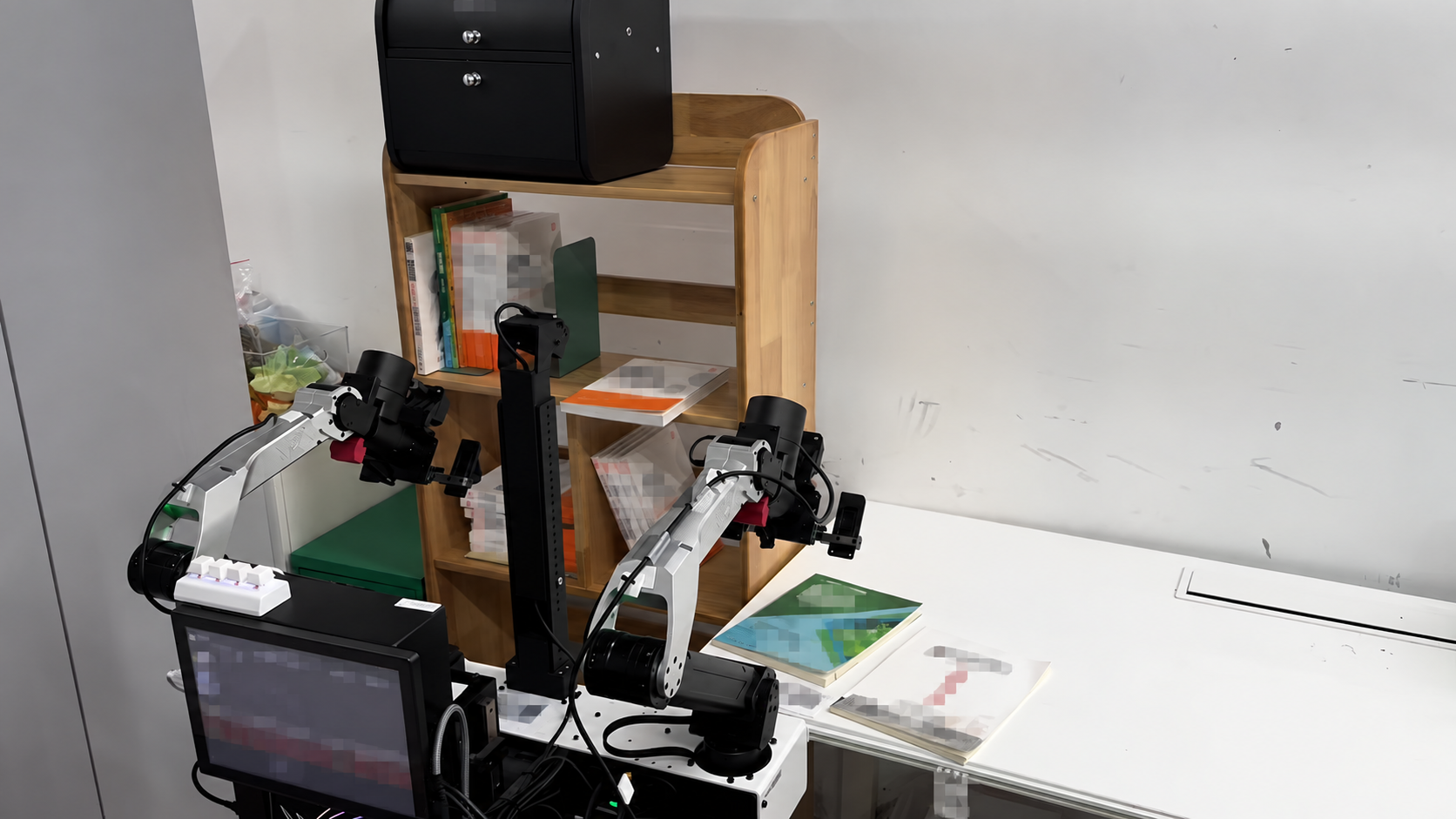}
  \caption{Arrange Bookshelf}
  \label{fig:realtasks:bookshelf}
\end{subfigure} \hfill
\begin{subfigure}[t]{0.32\textwidth}
  \includegraphics[width=\textwidth]{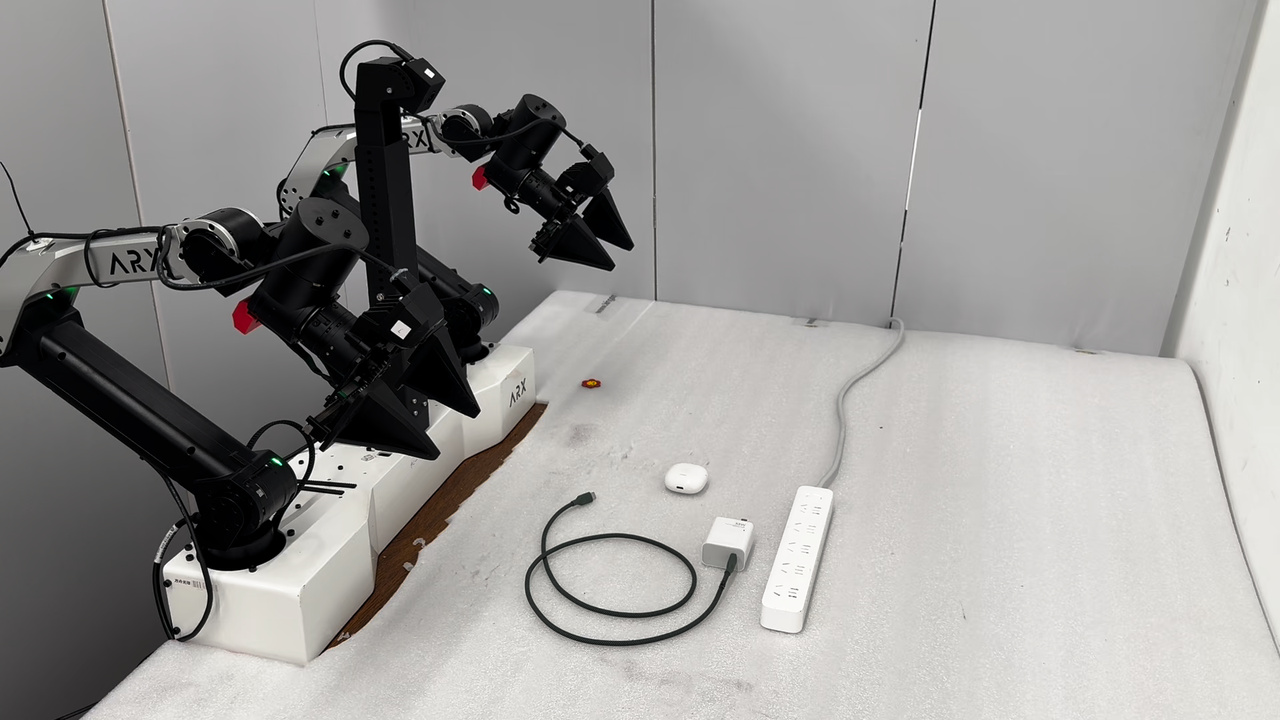}
  \caption{Charge Earphone}
  \label{fig:realtasks:earphone}
\end{subfigure} \hfill
\begin{subfigure}[t]{0.32\textwidth}
  \includegraphics[width=\textwidth]{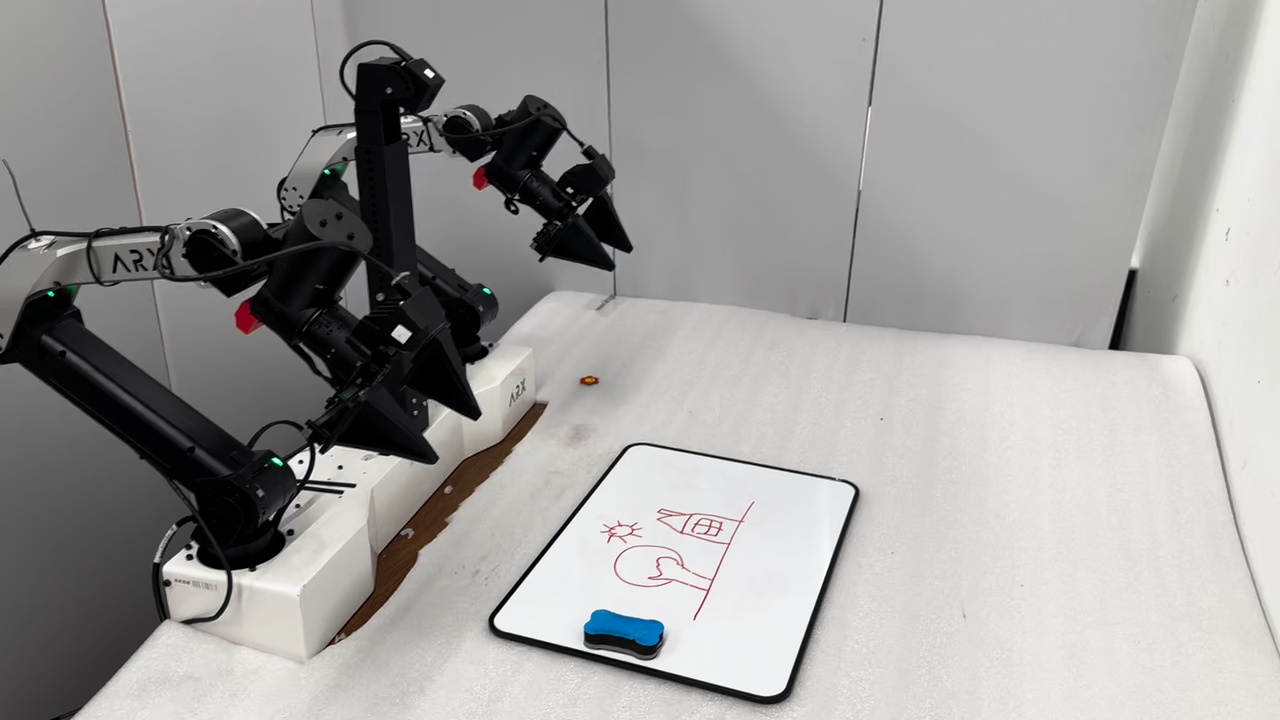}
  \caption{Clean Whiteboard}
  \label{fig:realtasks:whiteboard}
\end{subfigure}
\end{center}
\vspace{-0.1in}
\caption{The three real-robot tasks, each stressing a different temporal profile: (a) long-horizon mobile manipulation, (b) high-precision alignment, and (c) repetitive motion under a visual decision.}
\label{fig:realtasks}
\vspace{-0.1in}
\end{figure}

The curve is flat at the tight end and collapses steeply at the loose one. Already at the tightest setting, where little temporal compression is applied, most of the improvement over action chunking is in hand, so compression is not the source of the accuracy gains; the continuous smooth parameterisation is. The loose end is unambiguous: once the knots that resolve contact-rich phases are gone, accuracy falls below the action chunking baseline and nothing downstream recovers what the representation has discarded. The sweep therefore exhibits the trade the tolerance governs, compression rate against success rate, from which we select the deployed tolerance of each suite.

\paragraph{Executed length.}
\label{sec:exp:execlen} The second ablation holds the trained model fixed and involves no retraining: it changes only the row budget $H$, and with it how much of a prediction is committed. We sweep $H$ from short to long; since those rows decode to an adaptive number of steps, the executed span it yields is a mean rather than a setting. The middle and right panels of \Figref{fig:ablations} plot the result against both measures of what an executed length costs, the mean number of policy calls an episode requires and the mean number of control steps executed per call.

Matching the methods on inference cost gives a more direct and fairer account of what an adaptive executed length is worth, since a method that replans more often should be expected to score higher. On both suites the \method-Knot curve lies above the action chunking curve everywhere the two overlap, on either axis, and matching the \emph{baseline's} best accuracy requires $2.2$ times fewer policy calls per episode on RoboCasa and $2.5$ times fewer on LIBERO-Plus; Naive B-Spline sits between the two throughout. The two suites do disagree about the shape of the trade: on RoboCasa every curve is an inverted U, so committing too little of the prediction also hurts, whereas on LIBERO-Plus all three fall monotonically and the shortest budget is always best. How much of a window to commit is therefore a deployment-time choice, tuned per suite, while the span that committed window covers remains adaptive within it.

\subsection{Real-robot experiments}
\label{sec:exp:real}

We build three real-robot tasks, shown in \Figref{fig:realtasks}, chosen so that the temporal structure of the demonstrations differs as much as possible between them: \textbf{Arrange Bookshelf}, a long-horizon mobile manipulation task; \textbf{Charge Earphone}, a high-precision task whose two insertions demand tight alignment at the connector; and \textbf{Clean Whiteboard}, whose wiping strokes are highly repetitive and therefore highly compressible. Every method runs asynchronously with RTC. We score each trial against a set of annotated task nodes: progress (PG) is the fraction of nodes reached and success (SR) requires all of them; Appendix~\ref{app:real} describes the tasks, the nodes and the protocol in full.

\begin{table}[t]
\caption{Real-robot results over $30$ trials per task and method, all under asynchronous execution. Success rate (SR) requires every scored node of \Tabref{tab:real_nodes}; progress (PG) is the fraction reached, averaged over trials. Chunk is the mean control steps executed per policy call.}
\label{tab:real}
\begin{center}
\setlength{\tabcolsep}{4pt} \small
\begin{tabular*}{\textwidth}{@{\extracolsep{\fill}}lccccccccc@{}}
\toprule
 & \multicolumn{3}{c}{Arrange Bookshelf}
 & \multicolumn{3}{c}{Charge Earphone}
 & \multicolumn{3}{c}{Clean Whiteboard} \\
\cmidrule(lr){2-4}\cmidrule(lr){5-7}\cmidrule(l){8-10}
Model & SR (\%) & PG (\%) & Chunk & SR (\%) & PG (\%) & Chunk & SR (\%) & PG (\%) & Chunk \\
\midrule
Action Chunking & 20.0 & 52.2 & 32.0 & \textbf{26.7} & \textbf{73.3} & 32.0 & 50.0 & 67.5 & 32.0 \\
\method, Naive RTC & 10.0 & 35.6 & \textbf{39.8} & 13.3 & 43.3 & \textbf{46.5} & 46.7 & 68.3 & \textbf{53.3} \\
\method, JP-RTC & \textbf{23.3} & \textbf{54.4} & 38.6 & \textbf{26.7} & 57.5 & 45.7 & \textbf{60.0} & \textbf{74.2} & 52.4 \\
\bottomrule
\end{tabular*}
\end{center}
\vspace{-0.2in}
\end{table}

\paragraph{Results.} \method with JP-RTC exceeds the action chunking baseline on four of the six accuracy columns of \Tabref{tab:real} and matches it on a fifth. The largest margin is on Clean Whiteboard, the task designed to be the most compressible, at $10.0$ points of success rate and $6.7$ of progress; on Arrange Bookshelf it exceeds the baseline by $3.3$ and $2.2$ points. The one loss is instructive: on Charge Earphone the two reach the same success rate while the baseline is ahead on progress, and the gap sits at the socket insertion in the middle of the task, itself a fine manipulation stage. A continuous representation is the less reliable of the two at an insertion the demonstrations perform identically every time, a real cost on the one task built around precision, and it costs no completed trials. Meanwhile the baseline executes exactly $32$ control steps per call by construction while we decode $1.2$ to $1.6$ times as much from a parameter window of comparable size, so the efficiency gain costs no accuracy here, as in simulation.

The two RTC variants make the sharper point, being one model under two placements of the chunk continuity constraint. Naive RTC decodes the \emph{longer} chunks of the two on all three tasks and yet performs clearly worse on all three, in both success and progress, while what JP-RTC gives up to recover that is approximately one control step of chunk length. Constraining spline parameters directly therefore yields no gain in the length of the executed chunk, while degrading the quality of the actions within it.

\begin{figure}[t]
\begin{center}
\includegraphics[width=0.9\textwidth]{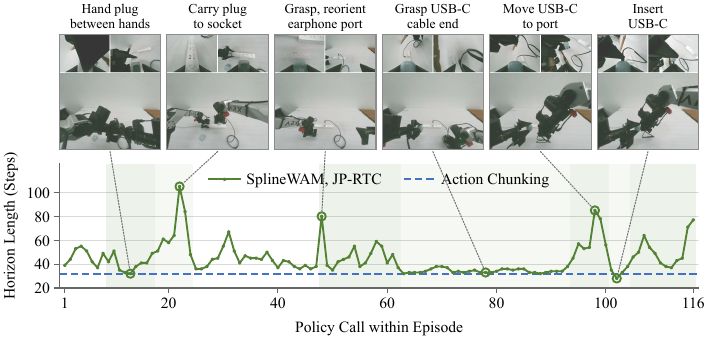}
\end{center}
\vspace{-0.1in}
\caption{Horizon length at every policy call of one successful Charge Earphone trial under \method with JP-RTC. Shaded bands mark the stages we identify from the recorded observations, with a keyframe above each stage's extremum.}
\label{fig:horizon_case}
\vspace{-0.2in}
\end{figure}

\paragraph{Qualitative analysis.} \Figref{fig:horizon_case} plots the horizon \method decodes at every policy call of one successful Charge Earphone trial. The horizon tracks the structure of the task rather than the clock, and it separates the stages into two groups. Free-space transfers, carrying the plug towards the socket and moving the USB-C connector towards the earphone port, are smooth enough for few knots to describe and decode into long chunks, several times the baseline's fixed horizon. Contact and alignment, closing the gripper on the cable end and the final insertion, attract knots densely and fall back to roughly the baseline's replanning rate, the shortest chunks of the episode arriving at the insertion itself. \method therefore replans at nearly the baseline's rate when precision demands it and stretches the horizon only where the motion is compressible, drawing its efficiency from the phases where a long chunk costs nothing, a trade a fixed horizon cannot make in either direction.

\section{Conclusion}

We propose \method, which replaces the fixed-length action chunk of a world action model with a fixed-size window of cubic B-spline knot times and control points. The window is fitted adaptively, so the temporal resolution of the prediction follows the complexity of the motion. Its knot times also carry the video supervision and the execution schedule, making the executed span and the replanning interval properties of the prediction rather than hyperparameters, and JP-RTC makes the representation usable asynchronously. On LIBERO-Plus and RoboCasa the method improves success rate by $8.2$ and $4.4$ points over an action chunking WAM while cutting policy calls per episode by $22\%$ and $26\%$. As policies grow past what fits on each robot and are served centrally, the call count sets how many robots a deployment can carry. The method carries over to three bimanual real-robot tasks under asynchronous execution. There it exceeds a fixed-chunk baseline on four of six accuracy measures and matches it on a fifth, while decoding $1.2$ to $1.6$ times as much executed motion per call. The decoded horizon lengthens through free-space transfers and contracts at insertion, without any explicit instruction to do so.

\bibliographystyle{iclr2027_conference}
\bibliography{iclr2027_conference}

\clearpage
\appendix
\section*{Appendix}

\startcontents[appendix]
\printcontents[appendix]{}{1}{\setcounter{tocdepth}{1}}
\clearpage

\section{Additional results}
\label{app:results}

This appendix collects the results deferred from \Secref{sec:experiments}: the per-category breakdowns behind the summary of \Tabref{tab:main}, the synchronous experiments on the original LIBERO suite, and the LIBERO-Plus counterpart of the asynchronous comparison reported on RoboCasa in the main text. Protocols, baselines and efficiency measures are exactly those of \Secref{sec:exp:setup}.

\paragraph{Per-category breakdowns.} \Tabref{tab:main_libero} and \Tabref{tab:main_robocasa} expand the two suites of \Tabref{tab:main} into their task categories. The category columns show that the gains of \method are not carried by one factor or family: on LIBERO-Plus, \method-Knot leads the action chunking baseline on six of the seven perturbations, with camera viewpoint and language instruction the largest margins at $11.0$ and $19.4$ points, and light conditions the single factor where it does not; on RoboCasa it leads on all four families. The two alignment variants are close in most columns and the summary margin of $1.2$ and $1.6$ points comes from a few: knot-grid alignment gains $7.5$ points on background textures and $4.2$ on turning tasks, while giving back $2.5$ points on camera viewpoint.

\begin{table}[h]
\caption{LIBERO-Plus success rate (\%) by action representation, at $\varepsilon = 0.01$ under synchronous execution, expanding \Tabref{tab:main}. Perturbation factors are background textures (BT), camera viewpoints (CV), language instructions (LI), light conditions (LC), objects layout (OL), initial states (IS) and sensor noise (SN); averages are episode-weighted.}
\label{tab:main_libero}
\begin{center}
\setlength{\tabcolsep}{3pt} \small
\begin{tabular*}{\textwidth}{@{\extracolsep{\fill}}lcccccccccc@{}}
\toprule
 & \multicolumn{8}{c}{Task Performance} & \multicolumn{2}{c}{Efficiency} \\
\cmidrule(lr){2-9}\cmidrule(l){10-11}
Action representation & BT & CV & LI & LC & OL
& IS & SN
& Average & Calls/episode & Steps/call \\
\midrule
Action Chunking & 61.1 & 43.5 & 70.8 & 87.4 & 75.9 & 66.6 & 63.8 & 66.3 & 33.6 & 8.00 \\
BEAST & 65.1 & 43.1 & 65.3 & 89.8 & 75.8 & 61.8 & 71.6 & 66.6 & 35.5 & 8.00 \\
Naive B-Spline w/o video & 55.6 & 32.9 & 71.0 & 85.1 & 66.7 & 53.7 & 61.0 & 60.0 & 38.7 & 7.92 \\
Naive B-Spline & 58.7 & 51.4 & 82.4 & 89.5 & 77.8 & 73.2 & 64.4 & 70.7 & 27.3 & 8.36 \\
\midrule
\method-Window & 54.7 & 57.0 & 87.8 & 84.2 & 79.1 & 73.7 & 74.3 & 73.3 & 27.1 & 8.39 \\
\method-Knot & 62.2 & 54.5 & 90.2 & 87.3 & 80.1 & 73.5 & 74.3 & \textbf{74.5} & \textbf{26.2} & \textbf{8.41} \\
\bottomrule
\end{tabular*}
\end{center}
\vspace{-0.2in}
\end{table}

\begin{table}[h]
\caption{RoboCasa success rate (\%) by action representation at $\varepsilon = 0.08$ under synchronous execution, expanding \Tabref{tab:main} under the same protocol and efficiency measures as \Tabref{tab:main_libero}. Columns are task families; averages are episode-weighted.}
\label{tab:main_robocasa}
\begin{center}
\setlength{\tabcolsep}{3pt} \small
\begin{tabular*}{\textwidth}{@{\extracolsep{\fill}}lccccccc@{}}
\toprule
 & \multicolumn{5}{c}{Task Performance} & \multicolumn{2}{c}{Efficiency} \\
\cmidrule(lr){2-6}\cmidrule(l){7-8}
Action representation & Pick-and-place & Open/close & Turn & Coffee
& Average & Calls/episode & Steps/call \\
\midrule
Action Chunking & 53.1 & 89.7 & 72.7 & 62.0 & 69.1 & 21.9 & 16.00 \\
BEAST & 44.1 & 88.7 & 71.0 & 59.7 & 65.0 & 24.6 & 16.00 \\
Naive B-Spline w/o video & 34.2 & 66.7 & 65.7 & 62.3 & 55.0 & 32.7 & 14.26 \\
Naive B-Spline & 52.8 & 91.8 & 69.6 & 69.3 & 69.5 & 16.9 & 17.98 \\
\midrule
\method-Window & 55.2 & 95.3 & 72.1 & 69.0 & 71.9 & 17.5 & 18.10 \\
\method-Knot & 56.1 & 94.7 & 76.3 & 71.3 & \textbf{73.5} & \textbf{16.3} & \textbf{18.18} \\
\bottomrule
\end{tabular*}
\end{center}
\vspace{-0.2in}
\end{table}

\paragraph{Original LIBERO.} \Tabref{tab:libero} reports the four standard LIBERO task suites under synchronous execution; we did not run the asynchronous comparison on this suite. We place these results here rather than in the main text because success rates on unperturbed LIBERO are high and tightly clustered, with every representation that is not actively broken landing within about a point and a half of the others and \method-Knot only $0.1$ points above the action chunking baseline, so the suite separates the comparisons far less than LIBERO-Plus does, and conclusions drawn from it would rest on differences smaller than single-run noise. Two observations survive that compression of the scale. Removing the auxiliary video term is the one change that clearly hurts ($90.5$ against $95.7$ for the same representation with the term), and the efficiency ordering matches the other suites, with \method-Knot issuing $17.2$ calls per episode against $19.8$ for action chunking.

The tolerance sweep is the one place where this suite disagrees with the perturbed ones: $\varepsilon = 0.005$ reaches $98.0$ against $96.4$ at $\varepsilon = 0.01$, so on unperturbed LIBERO the least compressed representation is the stronger one, whereas on RoboCasa and LIBERO-Plus a moderate tolerance is slightly better.

\begin{table}[h]
\caption{Original LIBERO success rate (\%) across the synchronous experiments of \Secref{sec:experiments}, with blocks matching \Tabref{tab:main_libero} and \Figref{fig:ablations}. Efficiency was instrumented for a subset of the representations only; blank entries were not measured.}
\label{tab:libero}
\begin{center}
\setlength{\tabcolsep}{4pt} \small
\begin{tabular*}{\textwidth}{@{\extracolsep{\fill}}lccccccc@{}}
\toprule
 & \multicolumn{5}{c}{Task Performance} & \multicolumn{2}{c}{Efficiency} \\
\cmidrule(lr){2-6}\cmidrule(l){7-8}
Setting & Spatial & Object & Goal & Long
& Average & Calls/episode & Steps/call \\
\midrule
\multicolumn{8}{l}{\emph{Action representation} (at $\varepsilon = 0.01$)} \\
\quad Action Chunking & 97.4 & 99.4 & 95.0 & 93.2 & 96.3 & 19.8 & 8.00 \\
\quad BEAST & 95.4 & 98.4 & 94.4 & 90.8 & 94.8 & 20.2 & 8.00 \\
\quad Naive B-Spline w/o video & 86.6 & 96.8 & 94.2 & 84.2 & 90.5 & 21.6 & 7.15 \\
\quad Naive B-Spline & 98.2 & 98.4 & 95.2 & 91.0 & 95.7 & 17.6 & 8.76 \\
\cmidrule(l){1-8}
\quad \method-Window & 98.2 & 98.4 & 96.2 & 91.0 & 96.0 & 17.5 & 8.78 \\
\quad \method-Knot & 99.0 & 99.0 & 96.6 & 91.0 & \textbf{96.4} & \textbf{17.2} & \textbf{8.82} \\
\midrule
\multicolumn{8}{l}{\emph{Fitting tolerance}} \\
\quad $\varepsilon = 0.005$ & 99.2 & 99.8 & 97.0 & 96.2 & \textbf{98.0} &  &  \\
\quad $\varepsilon = 0.01$ & 99.0 & 99.0 & 96.6 & 91.0 & 96.4 &  &  \\
\quad $\varepsilon = 0.025$ & 94.4 & 97.2 & 90.4 & 81.0 & 90.8 &  &  \\
\quad $\varepsilon = 0.05$ & 79.8 & 73.0 & 79.0 & 66.8 & 74.7 &  &  \\
\bottomrule
\end{tabular*}
\end{center}
\vspace{-0.2in}
\end{table}

\section{Real-robot tasks}
\label{app:real}

This appendix describes the three real-robot tasks of \Secref{sec:exp:real} and the protocol used to score them.

\paragraph{Arrange Bookshelf.} Books scattered across a desk and on the shelf must be placed upright and aligned in the shelf. Several consecutive placements, each moving between desk and shelf, make this a long-horizon mobile manipulation task.

\paragraph{Charge Earphone.} The robot inserts a charging plug into a wall socket and connects the earphone case to the cable. Both insertions demand tight alignment at the connector, making this a high-precision manipulation task.

\paragraph{Clean Whiteboard.} The robot erases writing from a whiteboard and returns the eraser to its holder. Judging which regions are still marked requires visual understanding, but the wiping strokes themselves are highly repetitive and therefore highly compressible.

\paragraph{Platform and action space.} Charge Earphone and Clean Whiteboard run on an ARX AC One, a bimanual platform of two $6$-DoF arms on a fixed base; Arrange Bookshelf runs on a Lift-2S, the same bimanual upper body on a $3$-DoF mobile base with an additional prismatic waist. Observations are three $240 \times 320$ RGB streams from a head camera and one camera per wrist, and the control loop runs at $30$\,Hz. An action is $d = 20$ dimensional on the two fixed-base tasks, comprising per arm $3$ end-effector translation, $6$ rotation in the two-column representation and $1$ gripper, and $d = 24$ on Arrange Bookshelf, which adds base velocity and the waist. The parameter window keeps the $36$ rows of the simulated setting, and the constrained prefix again covers $|S| = 8$ control steps. The representation is full-dimensional continuous cubic B-splines on the real robot as in simulation, gripper dimensions included.

\paragraph{Protocol.} Every method runs asynchronously, since a real deployment cannot pause the robot for inference. We score each trial against the annotated task nodes of \Tabref{tab:real_nodes}. Arrange Bookshelf is scored on the three placements it consists of, one node per book shelved, so each node is worth $33.3\%$. The other two tasks are scored on four nodes worth $25\%$ each, the stages on which the methods separate: every method reaches the earlier stages in nearly every trial, and the stages after the last node succeed or fail exactly with it. Progress (PG) is the fraction of a task's nodes that a trial reaches, averaged over trials, and success (SR) requires all of them. We run $10$ scene configurations per task and $3$ trials per configuration, giving $30$ trials per method per task; configurations vary the initial pose and the objects involved, such as the orientation of the earphone charging port and the colour of the marker to be erased.

\begin{table}[h]
\caption{The scored nodes of each task, in order. Progress credits an equal share per
node reached: $33.3\%$ for Arrange Bookshelf and $25\%$ for the other two.}
\label{tab:real_nodes}
\begin{center}
\setlength{\tabcolsep}{4pt} \small
\begin{tabular*}{\textwidth}{@{\extracolsep{\fill}}lllll@{}}
\toprule
Task & Node 1 & Node 2 & Node 3 & Node 4 \\
\midrule
Arrange Bookshelf & first book shelved & second book shelved
  & third book shelved & --- \\
Charge Earphone & grasp plug & plug into socket
  & grasp USB-C & insert USB-C \\
Clean Whiteboard & erase two thirds & nearly all erased
  & fully erased & eraser returned \\
\bottomrule
\end{tabular*}
\end{center}
\vspace{-0.1in}
\end{table}

\section{Asynchronous execution in simulation}
\label{app:rtcsim}

The real-robot comparison of \Secref{sec:exp:real} runs every method asynchronously. This appendix isolates the same comparison in simulation under a controlled inference delay, where the synchronous references are available alongside it. The delay is injected as a fixed $8$ control steps, matching the constrained prefix $|S| = 8$: the environment advances eight steps on the previous plan while the next is being computed, so every call arrives with exactly the overlap on which the continuity residual is defined. Fixing the delay in this way renders the three asynchronous rows comparable at a single operating point rather than averaged over a latency distribution, at the cost of not resolving how the gap scales with the delay, which Appendix~\ref{app:limitations} records as open.

\begin{table}[h]
\caption{RoboCasa success rate (\%) under synchronous and asynchronous execution with a controlled inference delay. The asynchronous rows sit below the synchronous references by construction, so the comparison of interest is among the three of them, which differ in where the continuity residual is defined.}
\label{tab:rtc_robocasa}
\begin{center}
\setlength{\tabcolsep}{8pt} \small
\begin{tabular*}{\textwidth}{@{\extracolsep{\fill}}lccccc@{}}
\toprule
Model \& inference setting & Pick-and-place & Open/close & Turn & Coffee
& Average \\
\midrule
Action Chunking, sync & 53.1 & 89.7 & 72.7 & 62.0 & 69.1 \\
\method, sync & 56.1 & 94.7 & 76.3 & 71.3 & \textbf{73.5} \\
\midrule
Action Chunking, async RTC & 42.6 & 88.8 & 70.4 & 65.3 & 65.1 \\
\method, async naive RTC & 40.5 & 88.3 & 70.3 & 61.3 & 63.8 \\
\method, async JP-RTC & 43.5 & 87.7 & 73.3 & 63.3 & \textbf{65.7} \\
\bottomrule
\end{tabular*}
\end{center}
\vspace{-0.1in}
\end{table}

Asynchronous execution trades accuracy for the removal of inference latency from the control loop. Real-time chunking is a method for keeping that trade acceptable, not for improving success rate, so the comparison that matters is among the three asynchronous rows of \Tabref{tab:rtc_robocasa}.

Applying ordinary real-time chunking directly in spline parameter space is actively harmful. \method with naive RTC lands $1.3$ points \emph{below} the action chunking baseline under the same asynchronous protocol: the representation that was the stronger of the two synchronously, by $4.4$ points, becomes the weaker one asynchronously. This is the failure mode predicted in \Secref{sec:inference} and drawn in \Figref{fig:jprtc}.

JP-RTC recovers this. It improves over naive RTC by $1.9$ points, bringing \method back above the action chunking baseline. Defining the residual on decoded actions and pulling it back through the decoder Jacobian is thus what makes a spline representation usable asynchronously at all: without it the representation is a liability under latency, and with it it is at least as good as a per-timestep chunk while retaining the temporal compression and the reduced call counts of \Tabref{tab:main}. A gap to synchronous execution remains, $7.8$ points, larger than the $4.0$ points that separate synchronous from asynchronous action chunking; Appendix~\ref{app:limitations} takes up what this leaves open.

\Tabref{tab:rtc_libero} is the LIBERO-Plus counterpart and reproduces the same conclusion. The two suites differ in the size of the asynchronous gap, which is far larger on LIBERO-Plus: JP-RTC reaches $47.5\%$ there against a synchronous $74.5\%$, a $27.0$-point drop where RoboCasa loses $7.8$. The ordering among the asynchronous rows is nonetheless the same, and the margin JP-RTC opens over naive spline-space RTC is wider, $8.1$ points against $1.9$; it also clears the action chunking baseline under the same protocol by $2.7$ points.

\begin{table}[h]
\caption{LIBERO-Plus success rate (\%) under synchronous and asynchronous execution, under the same protocol as \Tabref{tab:rtc_robocasa}. Columns follow \Tabref{tab:main_libero}.}
\label{tab:rtc_libero}
\begin{center}
\setlength{\tabcolsep}{6pt} \small
\begin{tabular*}{\textwidth}{@{\extracolsep{\fill}}lcccccccc@{}}
\toprule
Model \& inference setting & BT & CV & LI & LC & OL
& IS & SN & Average \\
\midrule
Action Chunking, sync & 61.1 & 43.5 & 70.8 & 87.4 & 75.9 & 66.6 & 63.8 & 66.3 \\
\method, sync & 62.2 & 54.5 & 90.2 & 87.3 & 80.1 & 73.5 & 74.3 & \textbf{74.5} \\
\midrule
Action Chunking, async RTC & 30.9 & 15.6 & 50.5 & 78.7 & 56.4 & 42.4 & 44.8 & 44.8 \\
\method, async naive RTC & 20.1 & 17.1 & 53.7 & 51.6 & 53.3 & 35.0 & 43.0 & 39.4 \\
\method, async JP-RTC & 25.7 & 27.9 & 65.5 & 61.7 & 59.7 & 39.3 & 50.5 & \textbf{47.5} \\
\bottomrule
\end{tabular*}
\end{center}
\vspace{-0.2in}
\end{table}

\section{Implementation details}
\label{app:impl}

Our codebase is derived from the public release of Fast-WAM~\citep{Yuan2026FastWAM}, which supplies the two-expert backbone, the Wan VAE~\citep{wan2025wan} video encoder, the flow-matching training loop and the evaluation harness. We keep these unchanged and add three components: the video--action alignment of \Secref{sec:training}, the B-spline action space that replaces the raw action chunk as the action expert's prediction target, and JP-RTC in the asynchronous inference path. The baselines in \Secref{sec:exp:setup} are the same codebase with the corresponding component disabled.

\paragraph{Backbone and training objective.} \method is obtained by fine-tuning the Wan2.2-5B video generation model~\citep{wan2025wan} under the two-expert arrangement of Fast-WAM, in which a video expert and an action expert are trained jointly. The observation frames of a sample are encoded to latents by the Wan VAE, and the video expert predicts a flow target on those latents while the action expert predicts one on the B-spline parameter window. Two flow targets are therefore predicted per sample and the losses are summed as $\mathcal{L} = \lambda_v \mathcal{L}_{\mathrm{video}} + \lambda_a \mathcal{L}_{\mathrm{action}}$ with $\lambda_v = \lambda_a = 1$, with per-sample scheduler weights applied to each term. The action loss is a mean squared error on the flow target in the model's normalised output space, computed over the whole parameter window, both parameter blocks being normalised to the common scale described below.

Adaptive knot fitting uses the FITPACK implementation shipped with SciPy~\citep{dierckx1995curve,virtanen2020scipy}: we call its least-squares spline routine with an explicit knot vector and drive the knot-insertion loop of \Secref{sec:bspline} ourselves, so that the stopping rule is the elementwise maximum of \Eqref{eq:emax} in normalised action units rather than SciPy's own smoothing criterion. Fitting is done offline as a dataset preprocessing step; the resulting parameter windows are cached alongside the demonstrations, so training sees no fitting cost.

\paragraph{Fixed-size local windows.} Following \citet{Han2026BSP}, the fixed-size window of \Secref{sec:bspline} is a contiguous run of knots and control points of the full-episode spline, extended with boundary support so that the curve remains evaluable across the interval the window is meant to cover, and we use $L_{\mathrm{BSP}} = L_{\text{video}} + 3$ rows for an observation window of $L_{\text{video}}$ frames. The $+3$ is a budget we chose rather than an identity derived from $k$, so we set out how the $36$ rows of our configuration are allocated. A row carries a knot time and a control point together, and the two are consumed differently. On the knot side the boundary is symmetric: with knots $u_0, \ldots, u_{35}$ the cubic basis is fully supported only on $[u_3,\, u_{32}]$, so three knots at each end, six in all, serve as boundary support and the remaining thirty knots, spanning twenty-nine intervals, carry the evaluable curve. On the control-point side it is not. A basis function $N_{j,3}$ is supported on $[u_j,\, u_{j+4}]$, so $c_0, c_1, c_2$ are active on the first evaluable spans and are ordinary parameters, whereas $c_{32}, \ldots, c_{35}$ are supported only at or beyond $u_{32}$ and contribute nothing to the decoded trajectory: the window has $32$ effective control points, matching the $32$-step chunk of the baselines.

The trailing four are therefore structural padding rather than free capacity, and we keep them so that every row has the same layout. They are still supervised, since the parameter-space flow-matching loss described above runs over the whole window, and their knot times are not inert: $u_{32}$ is the right endpoint of the evaluable interval and so sets $t_{\max}$ and the decoded length $M$, while $u_{33}, u_{34}, u_{35}$ are what the multiplicity trimming below acts on. Only the four tail control points are without effect. Each row stores one absolute knot time alongside one control point, so a row is $1 + d$ dimensional, $[\,t,\; c\,]$ with $c \in \mathbb{R}^{d}$, for actions of any dimension $d$ the embodiment provides, where $t$ is the knot time \emph{relative to the current observation}, measured in data frames: an absolute position on the local time axis, not an interval between neighbouring knots. Windows are anchored by assigning each observation to the spline span containing it and subtracting the observation frame index from the global knot times. Since boundary support consumes rows at both ends of the evaluable interval, the number of rows exceeds the number of effective knots, and we speak of a fixed-size \emph{parameter window} rather than a fixed number of tokens. $L_{\mathrm{BSP}}$ and $L_{\text{video}}$ count different quantities, parameter rows in the one case and observation frames in the other, and the relation between them is a budget chosen so that a window spans a comparable stretch of the episode to the observation it is anchored to. Adaptive fitting decouples the two: at a fixed $L_{\mathrm{BSP}}$ the duration those rows describe varies with the motion, which is the property the representation is built to provide.

\paragraph{Parameter scaling.} The two blocks of a row are not naturally on the same scale: control points inherit the $[-1,1]$ normalisation of the action space, whereas a knot time is a frame offset that can run to a hundred frames or more. Both are therefore mapped to a common range before they reach the model, knot times by an affine map fitted over the dataset to the interval the local windows occupy, so that the action loss weights a knot-time residual and a control-point residual comparably and the unbounded coordinate cannot dominate the gradient. The model predicts, and JP-RTC operates in, these normalised coordinates throughout; decoding inverts the map before evaluating \Eqref{eq:bspline}, and the knot times quoted in the paper are in frames.

\paragraph{Decoding safeguards.} A predicted window may not be a well-formed spline, so before evaluating one the decoder makes the knot times non-decreasing with a running maximum. Indexing the rows $j = 0, \ldots, L_{\mathrm{BSP}}-1$, it then restricts evaluation to $[\,u_k,\, u_{L_{\mathrm{BSP}}-1-k}\,]$, the interval over which the cubic basis is fully supported, so the $k$ knot times at each end serve as boundary support only. It also trims surplus repeated endpoint knots whose multiplicity would exceed what a cubic spline admits.

The running maximum requires a caveat, since it repairs at decode time what the parameterisation does not guarantee. Flow matching generates in an unconstrained $\mathbb{R}^{L_{\mathrm{BSP}} \times (1+d)}$ and cannot enforce $t_j \le t_{j+1}$, and the repair proceeds by setting an out-of-order knot equal to its predecessor rather than by reordering, so an inversion becomes a coincident interior knot, which reduces the curve from $C^2$ to $C^1$ at that point and further still if several knots coincide. In practice the fitted windows the model is trained on are monotone by construction and the predictions inherit this closely enough that the safeguard is seldom active, and the seam smoothing applied on the real robot removes what remains; the guarantee is nonetheless a decode-time clamp rather than a property of the representation. Predicting positive increments $\Delta u_j > 0$ and accumulating them would make monotonicity structural, at the cost of turning knot times into a quantity whose errors compound along the window, and whose absolute anchoring to the current observation would have to be recovered by summation. That anchoring is relied upon both by the alignment of \Secref{sec:training} and by the target shift of \Secref{sec:inference}. We compared the two parameterisations in preliminary experiments and found absolute times the better of the two, and we retain them for that reason.

\paragraph{Padding and masking.} Unlike a raw action chunk near an episode boundary, the action branch carries no padding mask: a tail window is completed by repeating the final control point, which is a well-formed spline rather than a placeholder, and every row is supervised. The video branch does carry one, because knot-grid alignment selects frames at knot offsets and a window near the end of an episode may request frames beyond it; repeated tail knots then select the same frame more than once and are marked as padding. Those positions are excluded from $\mathcal{L}_{\mathrm{video}}$, so that no gradient is taken against a duplicated target, and they are masked out of attention as keys, so that neither branch attends to them, while remaining present as queries in order to keep the latent grid at a fixed shape. Action parameters attend to the unpadded video positions under the backbone's mask, unchanged from Fast-WAM.

\paragraph{The gripper dimension.} The gripper command enters as one more continuous coordinate of the same spline, so a demonstration's open--close transition is a steep ramp rather than a step, and the decoded value is thresholded at the midpoint of the normalised range to recover a binary command, with no hysteresis. This has two consequences. Overshoot around a transition is harmless here, since the threshold discards it provided the sign is correct, and the low-level controller saturates the command in any case. The knot budget is affected more substantially: a transition in the gripper channel is precisely the kind of local feature to which the $\ell_\infty$ criterion of \Eqref{eq:emax} responds, and because all dimensions share one knot vector, knots inserted to resolve it are expended on the entire row. This is a genuine cost of the shared knot vector rather than a defect of the gripper's encoding, and it contributes to the decline in achievable compression as the action space widens, as Appendix~\ref{app:limitations} discusses. It is mitigated in practice by the fact that gripper transitions coincide with the contact phases that attract knots in any event.

\paragraph{Window sizes and action dimensions.} Observation windows use $L_{\text{video}} = 33$ frames with an action--video frequency ratio of $4$, giving $9$ supervised video frames and a $36$-row B-spline window against the $32$-step raw action chunk of the baselines. Both suites drive a single arm with $d = 7$ dimensional actions (3 translation, 3 axis-angle, 1 gripper), so a parameter row is $8$ dimensional and a window has $n = 288$ parameters; the constrained prefix of JP-RTC covers $|S| = 8$ control steps, giving $m_S = 56$. The representation is full-dimensional continuous cubic B-splines throughout; we do not use zero-order-hold gripper variants. At the row budgets of \Secref{sec:exp:setup} the baselines execute $8$ and $16$ steps of their $32$-step chunk under the same protocol.

\paragraph{Training and evaluation protocol.} The $24$ RoboCasa tasks of \Secref{sec:exp:setup} are grouped in our tables into four families: pick-and-place, opening and closing articulated objects, turning knobs and levers, and operating a coffee machine. Training uses programmatically generated trajectories and evaluation is on unseen layouts. On LIBERO we train for $10$ epochs at a global batch size of $128$. On RoboCasa we train for $4$ epochs at a global batch size of $256$, using $300$ demonstrations per task generated with MimicGen~\citep{mandlekar2023mimicgen}. Both use a peak learning rate of $10^{-4}$ with a decaying schedule, and every method in a given comparison is trained under the identical budget, sharing the demonstrations, the action normalisation and the low-level controller, so that a comparison changes only the action representation or the video--action alignment. Reported numbers are single-run point estimates.

The LIBERO-Plus results are zero-shot: we train no model on LIBERO-Plus. Each LIBERO-Plus number is produced by taking the model trained on the LIBERO training split and evaluating it under the seven perturbation factors without any adaptation, fine-tuning, or perturbation-specific model selection. The comparison in \Secref{sec:exp:main} is therefore a generalisation test rather than an in-distribution one, which is what makes the margins there larger than on the unperturbed suite: all representations fit the LIBERO training distribution well enough that success rates saturate, and they separate only once the observation is displaced.

\section{Derivation of the JP-RTC update}
\label{app:rtc}

This appendix derives \Eqref{eq:rawrtc} from the RTC update it generalises. With $A^\tau_t$ the action chunk at flow-matching time $\tau$, $Y$ the guidance target formed from the previous chunk, and $W$ the mask that weights the frozen prefix at $1$ and decays over the overlap region, \citet{Black2025RTC} correct the sampler's velocity by
\begin{equation} v_{\Pi\mathrm{GDM}}(A^\tau_t, o_t, \tau) = v(A^\tau_t, o_t, \tau) + \min\!\left(\beta, \frac{1-\tau}{\tau \cdot r_\tau^2}\right) \bigl(Y - \widehat{A}^1_t\bigr)^{\!\top} \operatorname{diag}(W) \frac{\partial \widehat{A}^1_t}{\partial A^\tau_t},
\label{eq:rtc}
\end{equation}
\begin{equation} \text{where}\quad \widehat{A}^1_t = A^\tau_t + (1-\tau) v(A^\tau_t, o_t, \tau), \qquad r_\tau^2 = \frac{(1-\tau)^2}{\tau^2 + (1-\tau)^2},
\label{eq:rtcaux}
\end{equation} and $\widehat{A}^1_t$ is the one-step estimate of the clean chunk, the weight being clipped at $\beta$ for stability at the small step counts used in control. Throughout, $\tau$ is the flow-matching time, $z^\tau_t$ the parameter window, $\widehat{z}^1_t$ its one-step estimate, $n = \dim z$, and $m_S = |S| \cdot d$ the number of constrained raw-action scalars ($m_S = 56$ against $n = 288$ in our configuration, though only a local subset of those $n$ parameters acts on the prefix, as the damping paragraph below makes precise). Integer control-step offsets are written $s$, reserving $\tau$ for the flow-matching time throughout. We derive the update under RTC's hard mask, so that $S$ is exactly the constrained set and $W_S = I$; the soft mask of \citet{Black2025RTC} enters at the end as a weighting of the residual. The guidance target is the previous parameter window $Y_z$, and the robot has committed to the segment of it that overlaps the arriving window, $y_S = D_{S + \Delta t}(Y_z)$ for an inference latency of $\Delta t$ control steps, the shift placing the two windows on a common time origin at the current observation; equivalently $y_S$ may be read off the controller's buffer of committed actions directly, which is what the implementation does. We write $e_S = y_S - D_S(\widehat{z}^1_t) \in \mathbb{R}^{m_S}$ for the unweighted residual and $r_S = W_S e_S$ for its masked counterpart, with $W_S = \operatorname{diag}(w_S) \otimes I_d$ lifting the per-step weights $w_S \in \mathbb{R}^{|S|}$ to the $m_S$ action scalars. Because the residual is formed after decoding, the mask indexes control steps rather than parameter rows and acts identically on all $d$ dimensions of a step, which is the first of the two differences from the RTC update; the second is the pullback itself. Everything below is in the model's normalised coordinates, as in our implementation: $z$ is the parameter window the sampler generates, $D$ maps it to actions in normalised units, and $J_S$ is the Jacobian in those coordinates.

\paragraph{The guidance term is already a damped pseudoinverse.} Pseudoinverse guidance~\citep{Song2023PiGDM} treats chunk continuity as an inverse problem $Y = \mathcal{H}(A^1_t) + \epsilon$ with $\epsilon \sim \mathcal{N}(0, \sigma_y^2 I)$, and approximates the posterior over clean samples by an isotropic Gaussian centred on the one-step estimate,
\begin{equation} p_\tau(A^1_t \mid A^\tau_t) \approx \mathcal{N}\bigl(\widehat{A}^1_t,\; r_\tau^2 I\bigr),
\label{eq:app:post}
\end{equation} with $r_\tau^2$ as in \Eqref{eq:rtcaux}. For a linear $\mathcal{H}(A) = HA$ the pushforward of \Eqref{eq:app:post} is again Gaussian, $\mathcal{N}(H\widehat{A}^1_t,\, r_\tau^2 H H^\top + \sigma_y^2 I)$, giving the closed-form guidance score
\begin{equation} \nabla_{A^\tau_t} \log p_\tau(Y \mid A^\tau_t) \approx \bigl(Y - H\widehat{A}^1_t\bigr)^{\!\top} \bigl(r_\tau^2 H H^\top + \sigma_y^2 I\bigr)^{-1} H \frac{\partial \widehat{A}^1_t}{\partial A^\tau_t}.
\label{eq:app:pigdm}
\end{equation} The matrix inverted in \Eqref{eq:app:pigdm} is a \emph{damped} Gram matrix: factoring out $r_\tau^{-2}$, which \Eqref{eq:rtc} already carries in its guidance weight, leaves the Tikhonov pseudoinverse $H^\top(HH^\top + \lambda I)^{-1}$ with
\begin{equation} \lambda = \sigma_y^2 / r_\tau^2 .
\label{eq:app:lambda}
\end{equation} The damped solve is therefore the exact form of the guidance term, with $\lambda \to 0$ encoding the belief $\sigma_y = 0$ that the target is matched exactly. This reading fixes the meaning of $\lambda$ but not its value, which we set from the spectrum of $J_S$ as \Eqref{eq:app:reldamp} describes.

\paragraph{Replacing the mask with a decoder.} For \method the generated variable is $z$ but the quantity that must match the previous chunk is the decoded prefix, so $\mathcal{H}(z) = D_S(z)$, which is nonlinear. \citet{Song2023PiGDM} handle nonlinear measurements by supplying a map $\mathcal{H}^\dagger$ with $\mathcal{H}(\mathcal{H}^\dagger(\mathcal{H}(z))) = \mathcal{H}(z)$, as they do for quantisation and JPEG, but $D_S$ admits no such closed form: inverting it means choosing both control points and the knot times that reparameterise them, and the least-squares fit that would do so is itself iterative. We therefore linearise at the posterior centre,
\begin{equation} D_S(\widehat{z}^1_t + \delta) = D_S(\widehat{z}^1_t) + J_S\delta + O(\|\delta\|^2),
\label{eq:app:lin}
\end{equation} under which the pushforward of \Eqref{eq:app:post} is again Gaussian, $\mathcal{N}(D_S(\widehat{z}^1_t),\, r_\tau^2 J_S J_S^\top + \sigma_y^2 I)$, so substituting $H \mapsto J_S$ into \Eqref{eq:app:pigdm} yields the bracketed update of \Eqref{eq:rawrtc},
\begin{equation} \delta^{\star} = J_S^\top \bigl(J_S J_S^\top + \lambda I\bigr)^{-1} e_S .
\label{eq:app:delta}
\end{equation} Equivalently $\delta^\star$ minimises $\|J_S\delta - e_S\|_2^2 + \lambda\|\delta\|_2^2$, the regularised least-squares correction that moves the decoded prefix towards the committed actions to first order. We use the dual form because it inverts an $m_S \times m_S$ matrix rather than an $n \times n$ one, the two being identical for $\lambda > 0$.

\Eqref{eq:app:lin} is to be read branch-wise: the decoding safeguards of Appendix~\ref{app:impl} are non-smooth where knots coincide or a knot is clamped, and the piecewise polynomial changes form as the active span switches, so the expansion holds where these discrete states are locally constant, the generic case since the non-differentiable set has measure zero in $z$; elsewhere the value automatic differentiation returns is implementation-dependent. The expansion is used only locally: $\delta^\star$ corrects the velocity rather than the sample, integrated over the denoising trajectory, so the decoded prefix is drawn towards $y_S$ across steps with the linearisation refreshed at each. The neglected term grows as $O(\|\delta\|_2^2)$ while the Tikhonov penalty acts on $\|\delta\|_2^2$, so the $\lambda$ that damps ill-conditioned directions also suppresses large steps, which bounds the update without certifying the first-order model at $\delta^\star$.

\paragraph{Locality and cost.} The Jacobian is local: cubic B-spline decoding is piecewise polynomial in the knot intervals, so \Eqref{eq:rawrtc} is one damped Gauss--Newton step per denoising step, not a projection onto an action-continuity constraint set. The extra cost over ordinary RTC is therefore one Jacobian and one $m_S \times m_S$ solve per denoising step, with $m_S = |S| \cdot d$ small.

This accounting rests on one approximation, left implicit in \Eqref{eq:rawrtc} and made explicit here. Written in full, the guidance term of \Eqref{eq:rtc} carries a factor $\partial \widehat{z}^1_t / \partial z^\tau_t$ relating the clean estimate to the current sample, and we take it to be the identity. Carried exactly, \Eqref{eq:rtcaux} gives $\partial \widehat{z}^1_t / \partial z^\tau_t = I + (1-\tau)\,\partial v / \partial z^\tau_t$, and the correction would require a vector--Jacobian product through the whole backbone at every denoising step, a cost comparable to one backward pass per step. This is not affordable in a control loop, and ordinary RTC does not incur it either. Dropping the term treats the one-step estimate as fixed for the purpose of the guidance, and is the same stop-gradient that \citet{Song2023PiGDM} rely on in practice. The claim of \Secref{sec:inference} that adaptive horizons cost no extra network evaluation is therefore exact, since the horizon is read off the predicted knot times. JP-RTC does add the Jacobian and the $m_S \times m_S$ solve described above, linear algebra that is negligible against a forward pass of a video-pretrained backbone but not, strictly speaking, free.

\paragraph{Reduction to ordinary RTC.} For a per-timestep chunk $D_S(z) = z_S$, so $J_S = P_S$ is a selection matrix with $P_S P_S^\top = I_{m_S}$, and \Eqref{eq:app:delta} becomes $(1+\lambda)^{-1} P_S^\top e_S$: the RTC update of \Eqref{eq:rtc} scaled by $(1+\lambda)^{-1}$, which is exactly $1$ as $\lambda \to 0$, and otherwise constant, since \Eqref{eq:app:reldamp} gives $\operatorname{tr}(J_S J_S^\top) = m_S$ and hence $\lambda = \lambda_{\mathrm{rel}}$; the factor then reparameterises $\beta$ rather than changing direction. This identifies what ordinary RTC assumes when applied to spline parameters, namely $J_S = I$: that perturbing a parameter row perturbs the correspondingly indexed action and nothing else, which is false for every cubic B-spline.

\paragraph{Setting the damping.} Local support can make $J_S$ ill-conditioned once knots are widely spaced. Split it into knot-time and control-point blocks, $J_S = [\,J_U \; J_C\,]$. A cubic basis function is supported on four knot spans, so constrained steps falling in $p$ consecutive spans that each contain at least one of them touch only $p + k$ control-point rows, giving $\operatorname{rank}(J_C) \le d\,(p+k) < m_S = d\,|S|$ whenever $p + k < |S|$, and then $J_C J_C^\top$ is singular. This makes precise the ambiguity to which \Secref{sec:inference} appeals, and identifies the freedom available to the pullback. The relevant dimension is not the full $n$ but the active dimension $n_{\mathrm{active}}$ of the rows whose support reaches $S$, whose control-point part the bound above caps at $d(p+k)$. In the smooth regime where an entire $|S| = 8$ step prefix falls inside one wide span, $p = 1$ admits at most $d(1+3) = 28$ control-point directions against $m_S = 56$ constrained scalars, the remaining $n - n_{\mathrm{active}}$ parameters exerting no influence on the prefix. The pullback is then solving an underdetermined system in which $\lambda$ selects the solution, and we therefore regard the damping as structural rather than as a device for absorbing residual numerical error. Writing $J_S J_S^\top = J_C J_C^\top + J_U J_U^\top$ with both terms positive semidefinite, Weyl's inequality gives
\begin{equation} \lambda_{\min}\bigl(J_S J_S^\top\bigr) \;\le\; \lambda_{\min}\bigl(J_C J_C^\top\bigr) + \|J_U\|_2^2 \;=\; \|J_U\|_2^2 .
\label{eq:app:weyl}
\end{equation} The knot-time block is itself weak, since $\partial a(s_i)/\partial u_j$ vanishes unless $u_j$ lies in the support of a basis function active at the constrained control step $s_i$, so for constrained steps interior to a wide span most of its entries are zero and the rest are small. A small $\|J_U\|_2$ then leaves the full Gram matrix with a small minimum eigenvalue, and both conditions hold in the smooth, widely spaced regime where continuity matters most. The spectrum is also not scale-free, since a unit perturbation of a knot time and of a control point move the decoded actions by different amounts, so an absolute $\lambda$ would mean different things at different knot densities. We set it relative to the spectrum,
\begin{equation} \lambda = \lambda_{\mathrm{rel}} \cdot \operatorname{tr}\bigl(J_S J_S^\top\bigr) / m_S ,
\label{eq:app:reldamp}
\end{equation} and use $\lambda_{\mathrm{rel}} = 10^{-1}$ for all reported results.

Two properties of \Eqref{eq:app:reldamp} should be stated explicitly, since neither follows from the Gaussian reading of \Eqref{eq:app:lambda}. First, it is held constant across $\tau$, whereas $\lambda(\tau) = \sigma_y^2 / r_\tau^2$ with a stationary $\sigma_y$ would diverge as $\tau \to 1$, since $r_\tau^2 \to 0$ there, and would thereby suppress the guidance of its own accord near the clean end. Holding $\lambda$ fixed instead amounts to an effective $\sigma_y^2(\tau) = \lambda\, r_\tau^2 = O((1-\tau)^2)$, a measurement precision that tightens as the sample is denoised, and it leaves the divergence of the prefactor $(1-\tau)/(\tau r_\tau^2)$ in \Eqref{eq:rawrtc} to be controlled by the clip $\beta$ rather than by the damping. We therefore regard \Eqref{eq:app:reldamp} as an empirical Gauss--Newton stabilisation whose scale is fixed by the spectrum of $J_S$, rather than as an exact implementation of the posterior score under stationary measurement noise. \Eqref{eq:app:lambda} establishes the dimensional meaning of $\lambda$, and a $\tau$-dependent schedule interpolating between the two is a natural variant that we did not find necessary.

Second, the penalty $\lambda\|\delta\|_2^2 = \lambda(\|\delta_U\|_2^2 + \|\delta_C\|_2^2)$ is isotropic over blocks that are not commensurable: $\delta_U$ displaces knot times, measured in frames, and $\delta_C$ displaces control points, measured in normalised action units, so a unit perturbation of each is penalised equally despite acting on the curve in different ways. Combined with $\|J_U\|_2 \ll \|J_C\|_2$ from the argument above, the minimum-norm solution places its correction almost entirely in $\delta_C$: JP-RTC reattaches a chunk by deforming the shape of the curve while leaving its timing essentially as predicted. This is the desired behaviour at a seam, since the knot times are what carry the horizon and a correction that rewrote them would trade continuity against the adaptivity the representation exists to provide. It is nonetheless a consequence of the scaling rather than a design choice, and a block-diagonal $\operatorname{diag}(\lambda_U I, \lambda_C I)$ would make the trade explicit and adjustable; we report the isotropic form.

\paragraph{Soft masking.} The derivation above uses the hard mask. RTC's soft mask replaces the binary weights by values decaying to zero over the overlap region, which we keep, applying it to the residual,
\begin{equation} \delta^{\star} = J_S^\top \bigl(J_S J_S^\top + \lambda I\bigr)^{-1} r_S , \qquad r_S = W_S\, e_S .
\label{eq:app:softdelta}
\end{equation} This step is a heuristic. Weights that differ across steps amount to a per-step measurement noise $\sigma_{y,i}^2 \propto W_{S,i}^{-1}$, for which \Eqref{eq:app:pigdm}, with the same $r_\tau^{-2}$ factored out as in \Eqref{eq:app:lambda}, gives
\begin{equation} \delta_{\Sigma} = J_S^\top \Bigl(J_S J_S^\top + \Sigma_y / r_\tau^2\Bigr)^{-1} e_S , \qquad \Sigma_y = \operatorname{diag}(\sigma_y^2) ,
\label{eq:app:sigmay}
\end{equation} on the same scale as \Eqref{eq:app:delta} and generalising its $\lambda$ to a diagonal matrix. Since $J_S J_S^\top$ couples constrained steps off-diagonally for a spline, \Eqref{eq:app:sigmay} differs from \Eqref{eq:app:softdelta} in direction and not only in weighting. Both cost one Cholesky factorisation; we prefer \Eqref{eq:app:softdelta} because it reduces to RTC's own soft-masked update when $J_S = P_S$ and requires no $\Sigma_y$ to be calibrated, the soft weights being an empirical device rather than a noise model we could set. We apply $W_S$ in raw-action space, before the pullback, so a weight sets how insistently a control step should match.

The cost of this choice should be stated precisely. \Eqref{eq:app:softdelta} is the solution of $\min_\delta \|J_S\delta - W_S e_S\|_2^2 + \lambda\|\delta\|_2^2$, which downweights the \emph{target} on a decaying step rather than the discrepancy at that step; the latter corresponds to the weighted least-squares form $\min_\delta \|W_S^{1/2}(J_S\delta - e_S)\|_2^2 + \lambda\|\delta\|_2^2$. For ordinary RTC the two agree in effect, since $J_S = P_S$ is orthogonal, the constrained steps decouple, and a weight tending to zero removes step $i$ from the problem entirely. For a spline they do not agree, because overlapping basis support renders $J_S J_S^\top$ off-diagonal: as $(W_S e_S)_i \to 0$ the target at step $i$ vanishes while $(J_S\delta)_i$ remains penalised at unit weight, so the solve is required to hold the decoded action at the far end of the mask near its unperturbed value rather than leaving it free. The decaying window is consequently stiffer at its tail than the weights suggest, and it is this stiffness that \Eqref{eq:app:sigmay} would relax. We retain \Eqref{eq:app:softdelta} for the reasons given above, and because the region it stiffens is the one the next call replaces, but a spline soft mask does not decay as cleanly as a per-timestep one.

\section{Limitations}
\label{app:limitations}

\paragraph{Precision at contact.} A cubic spline is smooth by construction, and the fit trades exactness for compression wherever the tolerance allows it. On the one real-robot task built around precision, \method reaches the baseline's success rate but not its progress (\Secref{sec:exp:real}), and the tolerance sweep shows accuracy collapsing as soon as the knots that resolve contact are removed (\Secref{sec:exp:ablations}). The representation is therefore best suited to tasks with a mix of free-space and contact-rich motion; for a task that is contact-rich throughout, the tolerance must be set tight enough that little compression remains.

\paragraph{Latency tolerance.} JP-RTC recovers the deficit that naive spline-space chunking opens, but a gap to synchronous execution remains on both simulated suites, and it is much larger on LIBERO-Plus than on RoboCasa (Appendix~\ref{app:rtcsim}). We do not claim that spline actions are as latency-tolerant as per-timestep chunks in general, only that JP-RTC is what makes them usable asynchronously at all. How the gap scales with the delay, and whether it can be closed by training with the asynchronous objective rather than correcting at inference time, we leave open.

\paragraph{One tolerance per dataset.} The fitting tolerance is a single scalar chosen per benchmark and held fixed for all of its tasks and all phases within a task. Adaptivity therefore operates within a window, through knot placement, but not across tasks: a suite containing both a purely repetitive task and a purely high-precision one would be served badly by any single value. Choosing $\varepsilon$ per task, or predicting it, is a natural extension we do not attempt.

\paragraph{Embodiment and action space.} A single knot vector describes all action dimensions at once, so the compression a given tolerance achieves is set by the hardest dimension to approximate, and the achievable compression falls as the action space grows. Our simulated suites drive a single arm with $d = 7$ dimensional actions, whereas the real-robot tasks are bimanual, with Arrange Bookshelf adding a mobile base. This is visible in \Tabref{tab:real}: Arrange Bookshelf decodes the shortest chunks of the three tasks despite involving the least fine manipulation, because its wider action space leaves less redundancy for one knot vector to absorb. A per-dimension or grouped knot vector would decouple this, at the cost of the single shared temporal structure the representation is built on, and we do not attempt it here. We also represent every dimension, including the binary gripper command, as a continuous cubic spline; we did not find this to require special handling, but embodiments with more discrete dimensions may.

\end{document}